\documentclass[journal]{IEEEtran}
\usepackage{times}
\usepackage{epsfig}
\usepackage{graphicx}
\usepackage{amsmath}
\usepackage{amssymb}
\usepackage{color}

\usepackage{booktabs}
\usepackage{multirow}

\usepackage{float}
\usepackage{subfigure}
\usepackage{amsfonts}
\usepackage{stfloats}
\usepackage{array}
\usepackage{booktabs}
\usepackage{fmtcount}
\usepackage{caption}
\usepackage{footnote}

\newcommand{\etal}{\textit{et al.}}
\newcommand{\ie}{\textit{i.e.}}
\newcommand{\eg}{\textit{e.g.}}

\newcommand{\wrt}{w.r.t.}

\ifCLASSINFOpdf
\else
\fi
\begin{document}
%
\title{Sharp Attention Network via Adaptive Sampling \\ for Person Re-identification}
%
%
%

\author{Chen~Shen,~Guo-Jun~Qi,~\IEEEmembership{Member,~IEEE,}~Rongxin~Jiang,~Zhongming~Jin,~Hongwei~Yong,~Yaowu Chen,\\~and~Xian-Sheng Hua,~\IEEEmembership{Fellow,~IEEE}%
\thanks{Copyright \copyright~2018 IEEE. Personal use of this material is permitted. However, permission to use this material for any other purposes must be obtained from the IEEE by sending an email to pubs-permissions@ieee.org.}
}

%
%

\markboth{IEEE Transactions on Circuits and Systems for Video Technology,~Vol.~XX, No.~XX,~XX~2018}%
{Shell \MakeLowercase{\textit{et al.}}: Bare Demo of IEEEtran.cls for IEEE Journals}
%



\maketitle

\begin{abstract}
	
In this paper, we present novel sharp attention networks by adaptively sampling feature maps from convolutional neural networks (CNNs) for person re-identification (re-ID) problem. Due to the introduction of sampling-based attention models, the proposed approach can adaptively generate sharper attention-aware feature masks. This greatly differs from the gating-based attention mechanism that relies soft gating functions to select the relevant features for person re-ID. In contrast, the proposed sampling-based attention mechanism allows us to effectively trim irrelevant features by enforcing the resultant feature masks to focus on the most discriminative features. It can produce sharper attentions that are more assertive in localizing subtle features relevant to re-identifying people across cameras. For this purpose, a differentiable Gumbel-Softmax sampler is employed to approximate the Bernoulli sampling to train the sharp attention networks. Extensive experimental evaluations demonstrate the superiority of this new sharp attention model for person re-ID over other related existing published state-of-the-art works
on three challenging benchmarks including CUHK03, Market-1501, and DukeMTMC-reID.

\end{abstract}


%
\IEEEpeerreviewmaketitle

\section{Introduction}
\label{sec:1}
%
%
%
%

 

\IEEEPARstart{W}{ith} the remarkable success of convolutional neural networks (CNNs)~\cite{krizhevsky2012imagenet, Simonyan15, szegedy2015going, He_2016_CVPR, Huang_2017_CVPR}, deep embedding methods, which aim to learn an end-to-end compact feature embedding of raw images, have made significant progress to advance many related computer vision tasks, including face verification~\cite{schroff2015facenet}, fine-grained image retrieval~\cite{wang2014learning, wei2017selective}, product search~\cite{bell2015learning} and person re-identification (re-ID)~\cite{li2014deepreid, yi2014deep}. Besides utilizing ``very deep" network structures and employing different kinds of loss functions (\eg,~Softmax~\cite{xiao2016learning, zheng2016person}, triplet~\cite{weinberger2009distance, hermans2017defense} and Online Instance Matching~\cite{Xiao_2017_CVPR}), a variety of solutions have been exploited intensively to enable more effective and efficient feature learning. Among these efforts is the attention mechanism focusing on the most discriminative parts of images in order to solve challenging recognition problems like person re-ID by distinguishing subtle fine-grained visual structures from other irrelevant parts~\cite{xu2015show, liu2017end, Wang_2017_CVPR, Zhao_2017_ICCV, rahimpour2017person}.

\begin{figure}[t]
	\centering
	\includegraphics[width=1.0\linewidth]{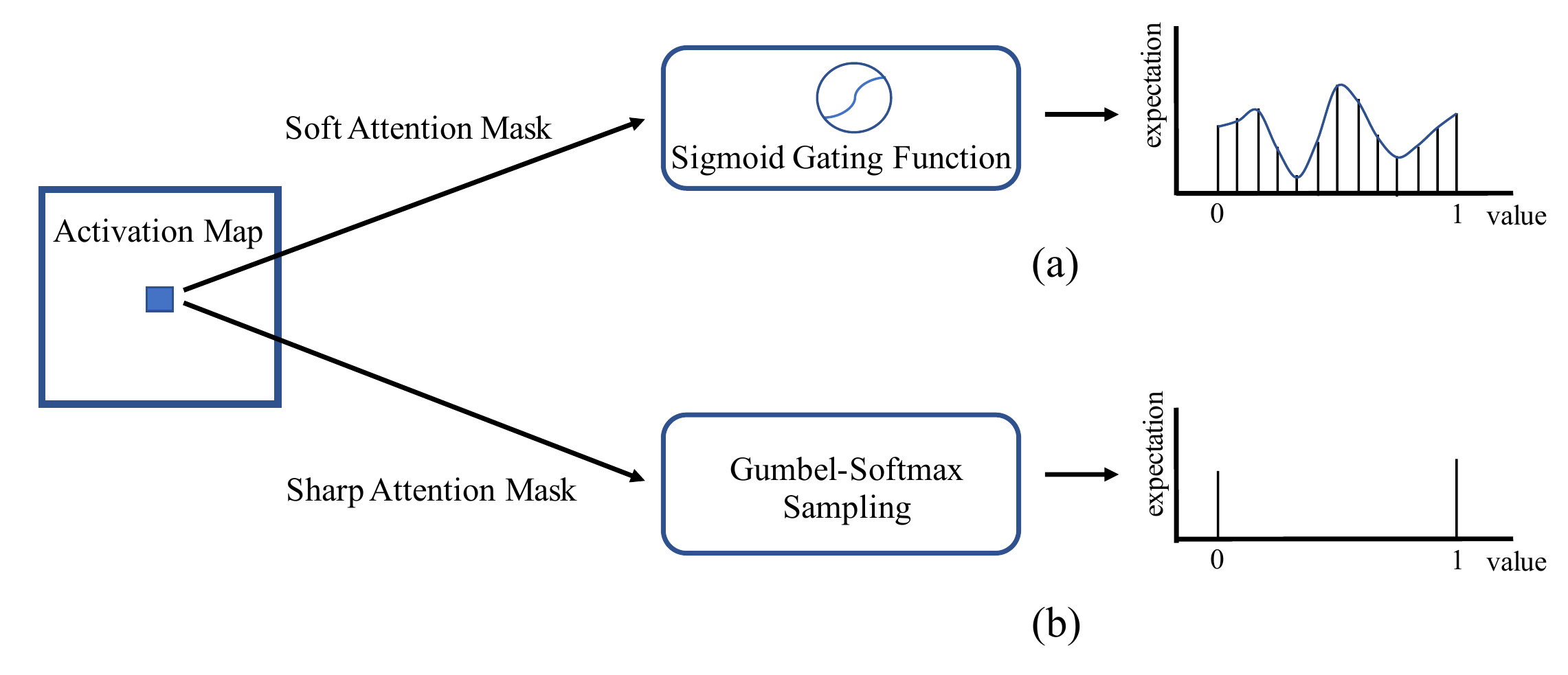}
	\caption{Two different types of attention mask generator. (a) Soft attention mask employed in~\cite{Wang_2017_CVPR, Zhao_2017_ICCV}. (b) Sharp attention mask introduced by us. }
	\label{fig:1}
\end{figure}

Existing attention mechanism in the context of deep learning often uses soft gating functions to select discriminative image parts. For example, recent works~\cite{Wang_2017_CVPR, Zhao_2017_ICCV} develop attention models incorporated into feed-forward CNNs that almost achieves state-of-the-art performance.
Their attention masks serve as control gates to perform an element-wise product with convolutional feature maps to localize discriminative visual structures. These attention masks are obtained from sigmoid functions with their ranges on $[0,1]$, as shown in Fig.~\ref{fig:1}(a). The continuous nature of these soft attention masks makes them have large uncertainty in localizing subtle discriminative parts for identifying different people and fine-grained object categories when their values are far from two assertive statuses of being attended ($1$) or unattended ($0$).

\begin{figure*}[t]
	\centering
	\includegraphics[width=0.9\linewidth]{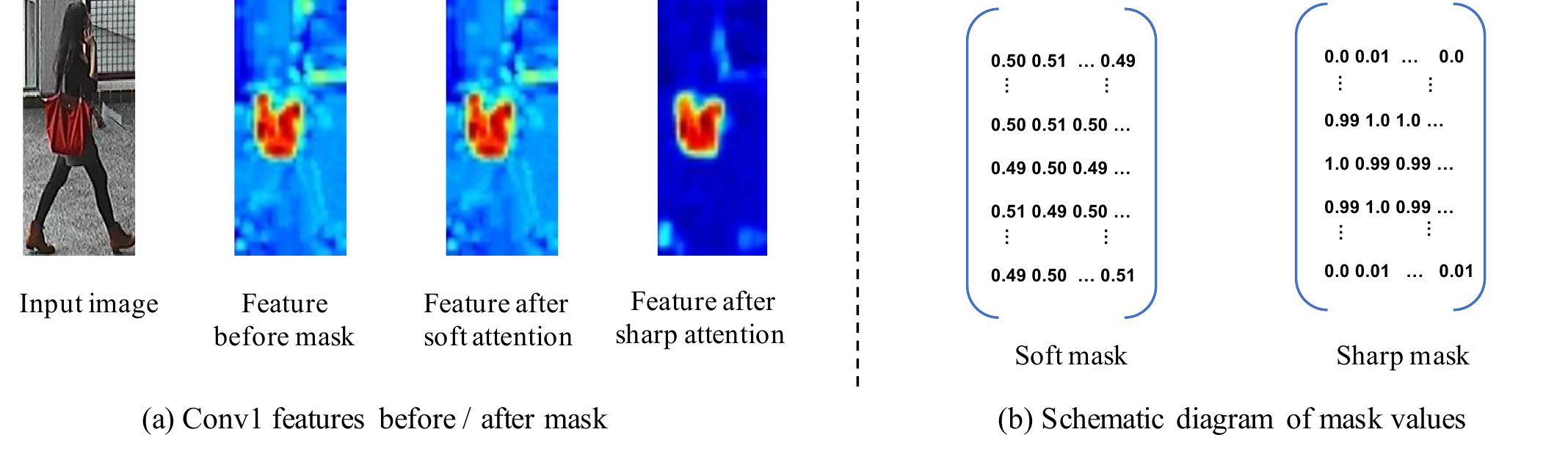}
	\caption{(a) The feature changing process after different masks. Compared with gating-based soft attention masks, the sampling-based sharp attention masks are more assertive in localizing subtle features relevant to re-identifying people (\eg,~a distinctive knapsack). (b) Schematic diagram of mask values. The sharp attention mask values are prone to be either $1$ (attended) or $0$ (unattended), with no attention ambiguity.}
	\label{fig:2}
\end{figure*}

For example, as shown in Fig.~\ref{fig:2}(a), the soft attention masks from sigmoid gates look ambiguous in selecting the most discriminative part (it is the knapsack in this case) for identifying the person in the given image. In many applications, we need sharper attention selectors that can more aggressively and assertively distinguish relevant visual structures from irrelevant ones. This is in particular important when the training examples are scarce, as otherwise the model could be prone to being overfitting to irrelevant visual structures.

The above challenge with gating-based soft attention mechanism inspires us to propose an alternative attention model, which can generate sharper attentions on these subtle visual structures that can identify different people and/or discriminate between fine-grained image categories. Unlike the gating-based model with soft uncertain attentions, we seek to generate sharper attention masks that are more assertive on selecting attended/unattended visual structures by directly sampling from the underlying feature maps.
As illustrated in Fig.~\ref{fig:2}, the sampled mask is much sharper than its soft attention counterpart -- it either attends ($1$) or unattends ($0$) to a particular location with no attention ambiguity. A sharper attention mask is particularly useful for person re-ID problem, which aims to retrieve and match the same person across non-overlapping surveillance camera views deployed at different locations. Fig.~\ref{fig:2} illustrates such a sharper mask is more sensitive to localize subtle details (\eg,~a distinctive knapsack) that can uniquely determine a particular person.

Technically, these sharper attention masks are generated through differentiable samplers drawn from Gumbel-Softmax distribution~\cite{jang2016categorical, maddison2016concrete} (see Fig.~\ref{fig:1}(b)). This distribution can separate the sampling randomness from the model parameters deciding where to select discriminative visual features. This allows us to backpropagate the error signals through these samplers of attention masks to update the trainable model parameters. Clearly, the discrete nature of these attention samplers ensures the generated attention masks could be sharper than soft attention masks with continuous values. Fig.~\ref{fig:1} summarizes the difference between the gating-based soft attention model and the proposed sharper attention model.

A cross-feature interaction learning scheme is also explored for enhancing the complementary benefit and joint learning compatibility of the original output features of the CNN backbone and the introduced sharp attention features, which further improves the re-ID performance. In addition, to achieve satisfactory results for those challenging re-ID scenarios where visual structures uniquely identifying a particular person can be localized only in a certain context (such as CUHK03 detected~\cite{li2014deepreid} and Market-1501~\cite{zheng2015scalable}), 
the sharp attention mask generator is essential to be equipped with a front-end unit, which can capture the high-level context-aware features in a larger receptive field to provide sampling guidance.
Full details of the above cross-feature interaction learning mechanism and context-aware sampling-guiding unit are to be presented in Sec.~\ref{sec:3}.

In summary, the contributions of this paper are threefold.
\begin{itemize}
	\item A novel sampling-based attention mechanism is proposed by training discrete attention masks from the CNN architectures in an end-to-end fashion.
	\item The generated attention is sharper than gating-based soft attention and can more assertively localize subtle visual structures to uniquely determine a particular person, which is exactly suitable for solving the person re-ID problem.
	\item The proposed sharp attention mask generator, cooperated with the well-designed cross-feature interaction learning scheme and the compact yet effective context-aware unit, achieves a consistent and significant performance gain compared with the baseline and other related methods on three challenging person re-ID datasets.
\end{itemize}

\section{Related Work}
\label{sec:2}

Deep learning based approaches have greatly boosted person re-ID task in recent years, as they incorporate feature extraction and distance metric into an unified framework, in which adaptive feature representations can be well learned under the supervision of a certain similarity metric.
Specifically, these methods utilize deep CNN architectures~\cite{krizhevsky2012imagenet, Simonyan15, szegedy2015going, He_2016_CVPR} to extract feature representations from raw images and employ different kinds of loss functions to optimize the embedding space, 
such that data points of positive pairs (\ie,~images from the same identity) are closer to each other than those of negative pairs (\ie,~images from different identities). 

Softmax loss, which regards the images of one identity as a category, is widely used recently and shows excellent superiority~\cite{xiao2016learning, zheng2016mars, zheng2016person}, as classification task can take full advantages of re-ID annotations and learn outstanding features with large inter-class variance. Xiao \etal~\cite{xiao2016learning} carefully design a baseline network where a Softmax loss is employed to optimize classification task and almost achieve state-of-the-art performance on some large datasets, \eg,~CUHK03~\cite{li2014deepreid}. 
On some other large datasets, the classification model also yields excellent performance without meticulous training sample selection~\cite{zheng2016mars, zheng2016person}.

Triplet loss~\cite{weinberger2009distance} and its variants~\cite{ding2015deep, oh2016deep, hermans2017defense, Chen_2017_CVPR} is another commonly employed loss function. An up-to-date work~\cite{hermans2017defense} shows that, using a variant of the triplet loss to perform end-to-end metric learning outperforms any other published method by a large margin. Chen \etal~\cite{Chen_2017_CVPR} design a generalized quadruplet loss, which can lead to the model output with a larger inter-class variation and a smaller intra-class variation compared to the triplet loss. Some other loss functions~\cite{huang2016local, zhou2017large, yao2017deep, Xiao_2017_CVPR, Zheng_2017_ICCV, Shen2017Deep} are also proposed for effective training. Online Instance Matching (OIM) loss is introduced by~\cite{Xiao_2017_CVPR}, which is scalable to datasets with numerous identities and converges much faster and better than the conventional Softmax loss. 
Shen \etal~\cite{Shen2017Deep} apply similarity perception loss to multi-level feature maps (\ie,~low-level and high-level). Therefore, the network can efficiently learn discriminative feature representations at different levels, which significantly improves the re-ID performance. 
A strong neural activation extraction scheme is proposed in~\cite{shen2017learning} to joint learn global features and local features.
In this paper, we follow the triplet mining strategy introduced by Hermans \etal~\cite{hermans2017defense} and adopt ResNet-50~\cite{He_2016_CVPR} network structure trained with triplet loss to produce a strong CNN baseline, which outperforms most of the existing deep learning frameworks.

Another clear trend for person re-ID is focusing on the feature extraction part and exploiting various techniques, which can be integrated into deep neural network with an end-to-end training pattern, for the purpose of more effective and efficient feature representing. Among these efforts, attention mechanism is one of the most recent architectural innovations. Liu \etal~\cite{liu2017end} first apply attention model to person re-ID problem. A recurrent soft attention based model is employed to generate different attention location information by comparing image pairs of persons through multiple glimpses and then integrate them together. However, RNN architecture and pairwise input are necessary in~\cite{liu2017end}, which is computationally expensive and intolerant for large-scale real-world applications. 

Later,~\cite{Zhao_2017_ICCV, rahimpour2017person, Wang_2017_CVPR} simplify the attention scheme to integrate into CNN structures. Zhao \etal~\cite{Zhao_2017_ICCV} exploit the Spatial Transformer Network~\cite{jaderberg2015spatial} as the hard attention model for searching discriminative parts given a pre-defined spatial constraint. Thus, a simple human part-aligned representation is proposed for handling the body part misalignment problem. Rahimpour \etal~\cite{rahimpour2017person} introduce a gradient-based visual attention model, which learns to focus selectively on parts of the input image for which the networks' output is most sensitive to. 
Wang \etal~\cite{Wang_2017_CVPR} present Residual Attention Network for image classification, built by stacking attention modules which generate attention-aware features, and bottom-up top-down feedforward structures which unfold the feedforward and feedback attention process into a single process. 
Importantly, a core component of the above methods is utilizing sigmoid function to regularize the mask values range to $[0, 1]$. In other words, they all generate soft attention masks. On the contrary, we address attention mask generation from another perspective, that is, generating sharper attention masks that are more assertive on selecting attended/unattended visual structures by directly sampling from the convolutional feature maps. The sharper attention mask is more sensitive to localize subtle details that can uniquely determine a typical person, which is particularly suitable for the person re-ID problem.

\section{The Proposed Approach}
\label{sec:3}

\begin{figure*}[t]
	\begin{center}
		\includegraphics[width=1.0\linewidth]{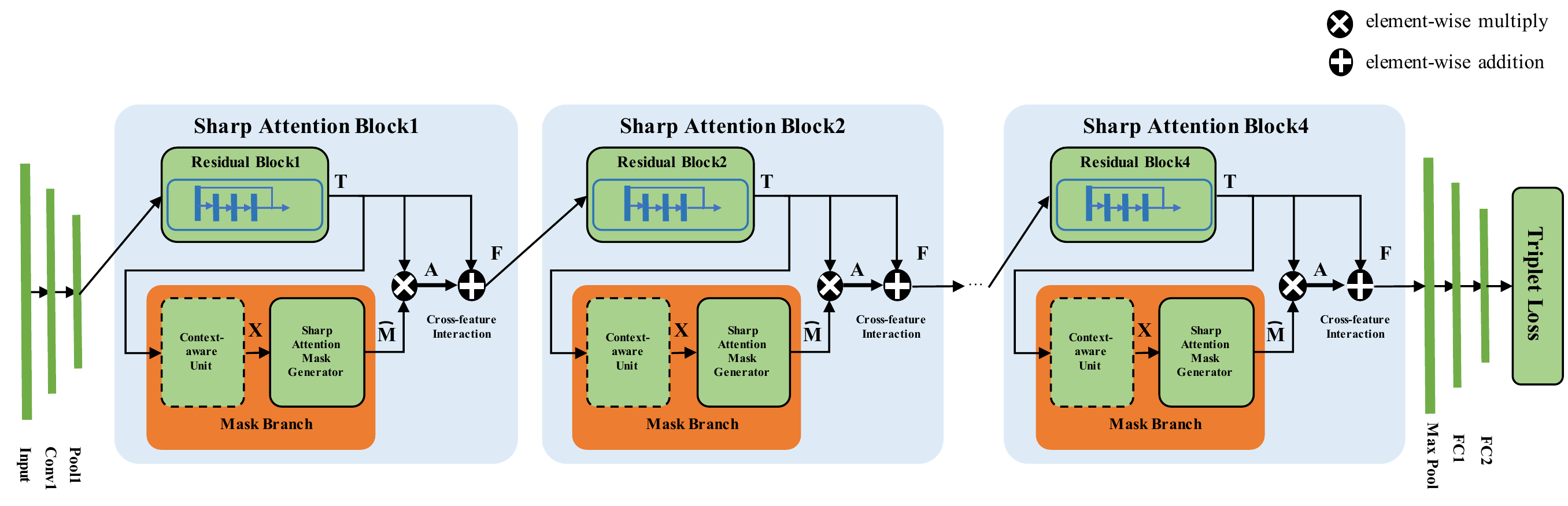}
	\end{center}
	\caption{Illustration of Sharp Attention Network structure. We adopt ResNet-50~\cite{He_2016_CVPR} as its backbone and each residual block can be expanded to a sharp attention block. For each attention block, $T$ represents the output of the trunk residual block, $\widehat M$ represents the attention mask through Gumbel-Softmax sampling, $A$ represents the attention-aware features, $F$ represents the final output of the sharp attention block, and $X$ represents the input of the attention generator. $X$ comes from either the output of the trunk residual block of ResNet-50 (\ie, $T$) or the output of the optional context-aware unit. We further introduce a cross-feature interaction learning scheme for maximizing the complementary benefit and compatibility of both the original feature and attention-aware feature representations.}
	\label{fig:3}
\end{figure*}

The proposed sampling-based sharp attention mechanism can be directly embedded into the state-of-the-art CNN frameworks. Fig.~\ref{fig:3} illustrates the overall architecture of a Sharp Attention Network along with its backbone CNN. In this paper, we adopt ResNet-50~\cite{He_2016_CVPR} as the backbone network\footnote{This choice is independent of our model design and others can be readily considered such as AlexNet~\cite{krizhevsky2012imagenet}, Inception~\cite{szegedy2015going} and VggNet~\cite{Simonyan15}.}. ResNet-50 is constructed by four sequential residual blocks and each of them can be expanded to a sharp attention block. 
In the CNN hierarchical framework, this block-wise sharp attention design naturally allows hierarchical multi-level (\ie,~from coarse level to fine level) attention learning to progressively refine the attention maps and boost the re-ID performance collaboratively.

Specifically, for each residual block, the original output features form its trunk, and after the last feature layer of the block is a mask branch which consists of an optional context-aware unit and a sharp attention mask generator. The former is a U-net~\cite{ronneberger2015u, noh2015learning} like structure to capture the high-level context-aware features in a larger receptive field to decide if an output feature should be selected by an attention mask. 
The attention mask generator employs Gumbel-Softmax sampling to acquire sharp attentions. This generator can be either based on the output of the backbone residual block or the output of the context-aware unit. 
Once an attention mask is generated, it is multiplied element-wise with the original trunk features to give attention-aware features. 
Additionally, we optimize the continuity of attention-aware features in the spatial domain by introducing total variation (TV) regularization penalty.

Conceptually, the above attention-aware feature learning aims at depicting the most discriminative local image regions of a person bounding box image, while the original trunk feature learning is dedicated to encoding the optimal global level features from the entire person image.	
In this sense, the attention-aware features can be viewed as some kind of local features and are largely complementary with the original features in functionality. Intuitively, their combination can integrate both advantages (\ie,~preserving global information and being more sensitive to particular local positions) and relieve the modeling burden from the same (particularly small) training data. Thus, we further introduce a cross-feature interaction learning scheme for maximizing the complementary benefit and compatibility of both the global and local feature representations. To be specific,
the original features are additively combined with the obtained attention-aware features by a skip connection.

Formally, suppose $T(x)$ is the output feature map of a trunk residual block with an input image $x$, and its size is $C \times H \times W$, where $C, H, W$ represent the number of elements in the channel, height and width dimensions, respectively. The mask branch generates an attention mask $M(x)$ of the same size with $T(x)$ through Gumbel-Softmax sampling (see the next subsection). The attention-aware feature map can be computed element-wise as
\begin{equation}\label{eq:1}  
A_{c,h,w}(x) = M_{c,h,w}(x) \times T_{c,h,w}(x), 
\end{equation}
\noindent where the subscript $(c, h, w)$ denotes the coordinate of an arbitrary position/pixel on the feature map and 
$c \in \{1, \ldots, C\}, h \in \{1, \ldots, H\}, w \in \{1, \ldots, W\}$ index the channel, the height and the width, respectively. 

After that, the final output $F(x)$ of the attention block additively combines the attention-aware features and the original residual block features as
\begin{equation}\label{eq:2} 
\begin{aligned}
F_{c,h,w}(x) &= A_{c,h,w}(x) + T_{c,h,w}(x) \\
&= (1 + M_{c,h,w}(x)) \times T_{c,h,w}(x). 
\end{aligned}
\end{equation}
By the above equation, we formulate the cross-feature interaction learning scheme for further enhancing the complementarity between original trunk features and attention-aware features, \ie,~perserving global characteristics while highlighting relevant local parts.

For the person re-ID problem, usually the resultant network is trained by minimizing the triplet loss~\cite{hermans2017defense}, denoted as $L_{tri}$. We will show that by using Gumbel-Softmax sampling to acquire $A$, the error signals can be backpropagated directly through the sampled $A$ to update the model parameters.

In the next four subsections, we will discuss in detail about four core components in the proposed networks: sharp attention mask generator, attention-aware feature continuity optimizing, cross-feature interaction learning and context-aware unit.

\subsection{Sharp Attention Mask Generator}
\label{sec:3_1}

We use the Gumbel-Softmax sampling to generate sharp attention masks, and it can be performed after either the output of the trunk residual block of ResNet-50 or the output of the optional context-aware unit (refer to Fig.~\ref{fig:4} in Sec.~\ref{sec:3_4}). 

Given an input $X$ to this sharp attention mask generator, it is first normalized onto an interval $[0,1]$ as
\begin{equation}\label{eq:3}
f(X_{c,h,w}) = \frac{X_{c,h,w} - \min_c}{\max_c - \min_c},
\end{equation}
\noindent where the subscript $(c, h, w)$ denotes the coordinate of an arbitrary position on the input;
$(h,w)$ ranges over all height and width locations and $c$ over all channels; $\max_c$ and $\min_c$ denote the maximum and the minimum value over $c$-th channel, respectively.
The normalized feature can be regarded as the probability of sampling this feature. Clearly, it tends to keep the highly activated features, while suppressing those weakly activated ones. It is indeed imposing a {\em parsimony} prior that pushes attention masks to only preserve the most relevant features while disregarding as many irrelevant ones as possible. Thus, it eventually leads to attention-aware features in which {\em strong gets stronger, and weak becomes weaker or even vanishes}.

Based on this probabilitic interpretation of normalized input $f(X_{c,h,w})$, a direct idea is to perform an in-place Bernoulli sampling according to it. However, the resultant attention-aware features would not be differentiable \wrt~$f(X_{c,h,w})$, and thus the back-propagation cannot be performed to update the network parameters through $X$. Fortunately, the Gumbel-Max trick~\cite{gumbel2012statistics, maddison2014sampling} provides an alternative way to draw an attention mask sample $M_{c,h,w}\in\{0,1\}$ from the Bernoulli distribution $\{\pi_1\triangleq f(X_{c,h,w}),\pi_0\triangleq 1-\pi_1\}$:
\begin{equation}\label{eq:4}
M_{c,h,w} = \mathop{\arg\max}_{j\in\{0,1\}}(g_j + \log\pi_j),
\end{equation}
\noindent where $g_0, g_1$ are i.i.d samples drawn from $\text{Gumbel}(0,1)$. After that, a Softmax function can be used to produce a continuous, differentiable approximation to relax $\arg\max$, which generates
\begin{equation}\label{eq:5}
\widehat M_{c,h,w}  = \frac{\exp((\log\pi_1 + g_1) / \tau)}{\sum_{j \in \{0,1\}} \exp((\log\pi_j + g_j) / \tau)}.  \ \ \ 
\end{equation}
\noindent When the temperature $\tau \to 0$, the above samples from the Gumbel-Softmax distribution~\cite{jang2016categorical, maddison2016concrete} become identical to those from the Bernoulli distribution.
Thus, we will employ the Softmax approximation $\widehat M_{c,h,w}$ as the attention masks applied to original residual block features to obtain attention-ware features. 

The Gumbel-Softmax distribution is smooth for $\tau > 0$, and thus we can compute the gradient $\partial \widehat M_{c,h,w} / \partial \pi_1$ as $\pi_1$ is separate from the random discrete samples $g_1$ and $g_0$ drawn from $\text{Gumbel}(0,1)$. 
In the implementation, we will start at a high temperature ($\tau = 1$) and gradually anneal to a small one ($\tau = 0.5$)~\cite{jang2016categorical}.

\subsection{Attention-aware Feature Continuity Optimizing}
\label{sec:3_2}

The proposed sampling-based sharp attention selectors can assertively localize subtle discriminative parts and eliminate irrelevant features. Nevertheless, because of the inevitable sampling randomness, a fraction of noisy masks (or features) occur in the spatial domain. For instance, a mask with value of $0$ (corresponding to unattended feature) appears among a bunch of masks with value of $1$ (corresponding to attended feature). Hence, we introduce total variation (TV) regularization penalty to optimize the continuity of attention-aware features, aiming to trim the above noisy and meaningless features, in the spatial domain. TV regularization~\cite{rudin1992nonlinear} is based on the principle that signals with excessive and possibly spurious detail have high total variation, and is remarkably effective at simultaneously preserving important details whilst smoothing away noise.

Technically speaking, given an arbitrary attention-aware feature $A \in \mathbb{R}^{C \times H \times W}$ where $C, H, W$ denote the number of pixel in the channel, height and width dimensions, we firstly introduce two matrices, denoted as:
$$
D_1 = 
\left[
\begin{matrix}
\begin{array}{rrrrr}
1 & -1 & \cdots & 0 & 0 \\
0 &  1 &  -1    & \cdots & 0 \\
\vdots & \vdots & \ddots & \ddots & \vdots \\
0 & 0  & \cdots & 1 & -1  \\
\end{array}
\end{matrix}
\right] \in \mathbb{R}^{(H - 1)\times H},
$$ 
and
$$
D_2 = 
\left[
\begin{matrix}
\begin{array}{rrrr}
1 & 0 & \cdots & 0 \\
-1 &  1 & \cdots & 0 \\
0 & -1 & \ddots & \vdots \\
\vdots & \vdots & \ddots & 1 \\
0 & 0 & \cdots & -1 \\
\end{array}
\end{matrix}
\right] \in \mathbb{R}^{W \times (W - 1)}.
$$
After that, the TV regularization penalty for $A$ is defined as:
\begin{equation}\label{eq:6}
L_{TV}^{A} = \sum_{c \in \{1,  \ldots, C\}} \|D_1 A_c \|_2^2 + \|A_c D_2 \|_2^2,
\end{equation}
where $A_c \in \mathbb{R}^{H \times W}$ represents the c-th channel feature map of $A$. And the gradient can be effectively computed as
\begin{equation}\label{eq:7}
\partial L_{TV}^{A} / \partial A_c = 2(D_1^\mathrm{T} D_1 A_c + A_c D_2 D_2^\mathrm{T}). 
\end{equation}
\noindent At last, the whole loss function is defined as:
\begin{equation}\label{eq:8}
L = L_{tri} + \mu \sum_{i} L_{TV}^{A_{i}}, 
\end{equation}
where $i$ indexes the sharp attention blocks (\ie, block1 to block4) and the hyperparameter $\mu$ is used to control the balance between the TV regularization penalty $L_{TV}$ and the aforementioned triplet loss $L_{tri}$.

\subsection{Cross-Feature Interaction Learning}
\label{sec:3_3}

For a typical sharp attention block, given the original residual trunk features (\ie,~global-level features) and the attention-aware features (\ie,~part-level features) above, we further consider a cross-feature interaction mechanism for enriching their complementary benefit and joint learning compatibility. Specifically, the cross-feature interaction learning scheme is formulated by Eq.~\ref{eq:2}  and can be implemented by an identity shortcut connection and element-wise addition (channel by channel).

We also tested concatenation of both attention-aware features and the original ResNet-50 residual block features, but found the resultant network is hard to converge. This shows that an additive combination is a better choice. In this way, the attention-aware features can also be viewed as attention residuals that can be additively combined with the original ResNet-50 features (see Fig.~\ref{fig:3}), similar to ideas in residual learning~\cite{He_2016_CVPR}, which can make it easier to optimize and gain performance improvement consistently.

\subsection{Context-aware Unit for Sampling Guiding}
\label{sec:3_4}
 
In the proposed framework, the attention masks can be generated directly by sampling the original residual block without involving any new network layers. This results in a {\em lazy} context-free sharp attention mask generator, which can still yield satisfactory performance improvement when we do not need to localize which image parts should be attended in a suitable context.

However, if visual structures that can uniquely identify a particular person can be localized only in a certain context, the attention mask generator should be equipped with a front-end unit to model visual contexts. For instance, in practical re-id scenarios, person images are typically automatically detected for scaling up to large visual data. Under these circumstances, to select a discriminative visual structure or not should rely on the high-level context-aware features in a larger receptive field to alleviate the negative impacts from background clutter, occlusion and missing body parts caused by the poor detected bounding boxes.  
Inspired by the ``U-net" like structure~\cite{ronneberger2015u, noh2015learning} in segmentation, detection and pose estimation, we introduce a context-aware unit. It is constructed by stacking convolutional layers and mirrored deconvolutional layers together. This structure can be viewed as a bottom-up forward and top-down feedback pipeline, which combines multi-scale visual information at various levels.

\begin{figure}[t]
	\begin{center}
		\includegraphics[width=0.9\linewidth]{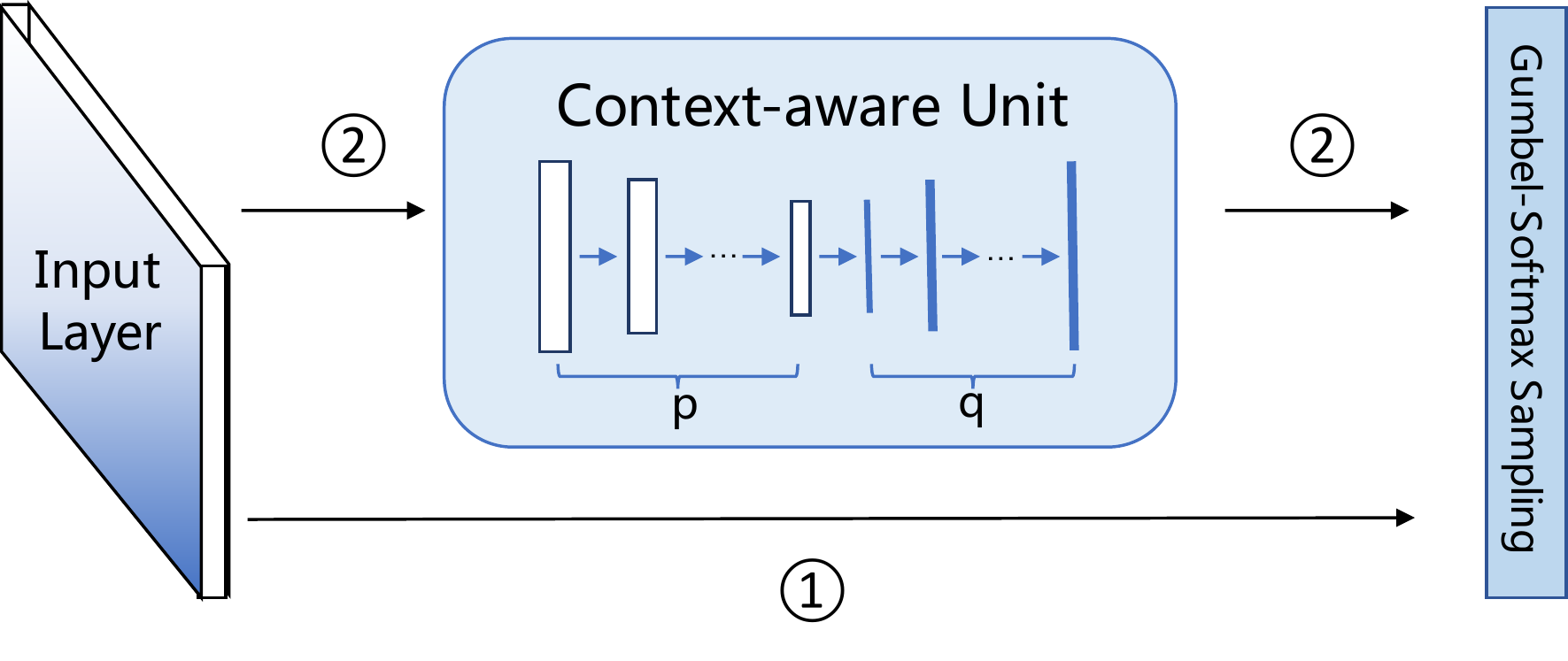}
	\end{center}
	\caption{Illustration of the context-aware unit structure. The input layer represents the last output layer of original residual block. \textcircled{\scriptsize 1} and \textcircled{\scriptsize 2} denote directly sampling and sampling through context-aware guiding, respectively. $p, q$ are two hyper-parameters denoting the numbers of convolutional residual units and deconvolutional layers within context-aware unit. In our experiments, we use the following setting: ${p = 1, q = 1}$.}
	\label{fig:4}
\end{figure}

For the convolutional layers, we reuse the architecture of the backbone residual units (with $2$ stride) but train it with a different group of parameters. On the other hand, in the deconvolutional layers, we simply employ a kernel filter of size $1 \times 1$ and a fractional $1/2$ stride to upsample the contextual feature maps. Eventually, the output layer of this ``context-aware network unit" has the same size as the input ResNet-50 features, but captures a larger size of receptive field and deeper context-aware features. In this fashion, attention masks can be generated side-by-side by sampling such contextual layer.
The structure of context-aware unit is illustrated in Fig.~\ref{fig:4}.

Notice that we just need a relatively compact yet effective structure (\ie,~a residual unit for down sampling and a deconvolutional layer for up sampling), cooperated with the proposed sampling-based attention mechanism, to achieve satisfactory performances. That is attribute to the superiority of sharper attention selectors over soft attention ones, which can more aggressively and assertively distinguish relevant visual parts. Conversely, some soft attention models (\eg,~\cite{Wang_2017_CVPR}) also consider context-aware information to guide attention selection but with complicated bottom-up top-down structure. This sub-network design with high complexity~\cite{Wang_2017_CVPR} is ineffective in model deployment and prone to being overfitting when only a small set of labeled data is available for model training.

\section{Experiments}
\label{sec:4}

\subsection{Datasets and Evaluation Protocols}
\label{sec:4_1}

We conduct experiments on three person re-ID datasets CUHK03~\cite{li2014deepreid}, Market-1501~\cite{zheng2015scalable} and DukeMTMC-reID~\cite{ristani2016MTMC, Zheng_2017_ICCV} widely used in literature. In our experiments, given a test probe image $I^p$ from one camera view and a set of test gallery images ${I_i^g}$ from other non-overlapping camera views, we first compute their corresponding deep feature representations by forward-feeding the images to a trained sharp attention network model, denoted as $x^p$ and $x_i^g$, then we compute the similarities (based on the Euclidean distance) between $x^p$ and $x_i^g$. After that, the ranked gallery list are returned in a descent order of the similarities.
The performances are evaluated by the commonly used Cumulative Matching Characteristics (CMC)~\cite{moon2001computational} top-k accuracy, which is an estimate of the expectation of finding the correct match in the top k retrieved items. Following~\cite{zheng2015scalable}, we also report the mean Average Precision (mAP) over all three datasets. All the experiments are performed in a single-query setting (\ie,~one query each time).

\textbf{CUHK03.}
The CUHK03 dataset consists of five pairs of camera views, including $14,097$ images of $1,467$ pedestrians. Each identity only appears in two disjoint camera views on the CUHK campus, on average with $4.8$ images in each view. Li \etal~\cite{li2014deepreid} provide two types of bounding boxes: labeled (human annotated) and detected (automatically produced by the DPM detector~\cite{felzenszwalb2010object}).
In this paper, we conduct experiments on both sets, using the provided training/testing splits ($1,267$ identities for training, $100$ for validation, and the last $100$ for testing).
For each test identity, two images are randomly sampled from different camera views as the probe and gallery images, respectively, and the average performance over 20 times is reported as the final result.

\begin{figure*}[t]
	\begin{center}
		\includegraphics[width=1.0\linewidth]{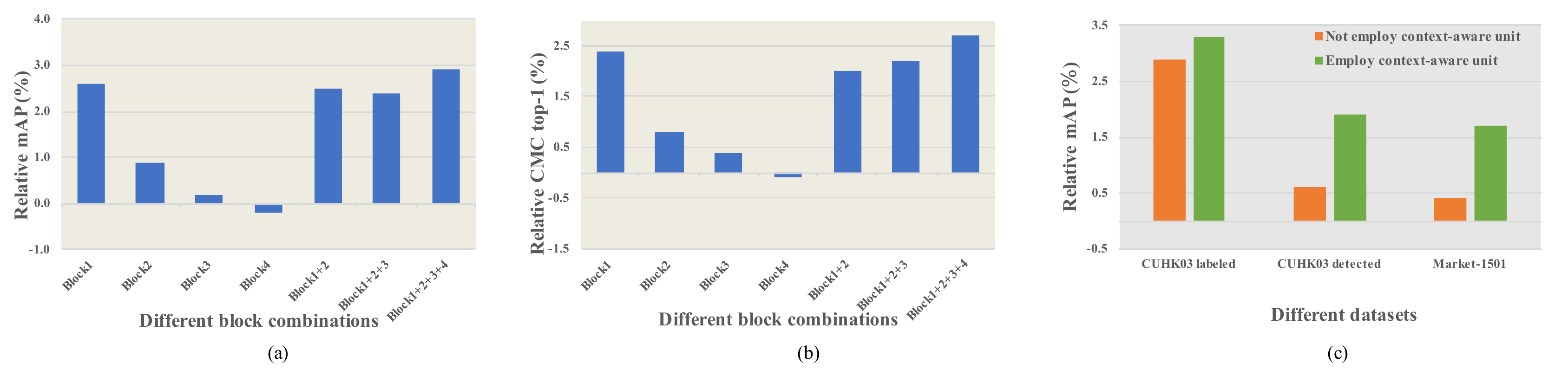}
	\end{center}
	\caption{(a), (b) The relative performances in terms of mAP and CMC top-1 accuracy over the baseline ResNet-50 with different block combination strategies on CUHK03 labeled, respectively. (c) The relative performances in terms of mAP over the baseline ResNet-50 with different sampling strategies on several datasets.}
	\label{fig:5}
\end{figure*}

\textbf{Market-1501.}
The Market-1501 dataset contains $32,668$ pedestrian images of $1,501$ identities captured from six cameras in different resolutions: five high-resolution cameras, and one low-resolution camera. It is a large-scale benchmark dataset for person re-ID. 
$19,732$ images are used for testing ($751$ identities, along with $2,793$ distractor images) and the remaining $12,936$ images ($750$ identities) are used for training.
There is an average of $17.2$ training images per identity in this set.
The people in the images are automatically detected by the deformable part model (DPM)~\cite{felzenszwalb2010object}, so the incorrect detections of people are common, along with partial occlusion, which makes the dataset challenging and close to real-world scenarios.

\textbf{DukeMTMC-reID.} DukeMTMC~\cite{ristani2016MTMC} is a newly-released multi-target, multi-camera pedestrian tracking dataset. It contains eight 85-minute high-resolution videos from eight different cameras. Hand-drawn pedestrian bounding boxes are available.
In this paper, we use its re-ID version benchmarked DukeMTMC-reID~\cite{Zheng_2017_ICCV}, which is a subset of the original dataset and contains $1,404$ identities appearing in more than two cameras. $702$ identities are selected into the training set and the remaining $702$ identities into the testing set. This results in $16,522$ training images, $2,228$ queries, and $17,661$ gallery images. In the testing set, one query image for each identity in each camera is picked and the remaining images are viewed as the gallery.

\subsection{Implement Details}
\label{sec:4_2}

\textbf{Network architecture.} We adopt ResNet-50~\cite{He_2016_CVPR}
as the backbone network, and replace the last $1000$-dimensional fully-connected layer with two fully-connected layers. The output features of the first FC layer are $1024$-dimension, followed by batch normalization~\cite{ioffe2015batch} and ReLU~\cite{krizhevsky2012imagenet} layers; the output of the second FC layer goes down to $128$-dimension, yielding the final feature representation. Same backbone architecture is adopted for all experiments for fair comparison.
During training, each input image is cropped from a random portion of the image sampled in $[0.64, 1.0]$ with an aspect ratio randomly chosen in $[2, 3]$. The cropped area is then resized to $256 \times 128$, for data augmentation. Horizontal flip and per-pixel mean subtraction are also used. During testing, for comparison we adopt the standard 10-crop testing~\cite{krizhevsky2012imagenet}. 
In experiments, which residual block of ResNet-50 should be extended to a sharp attention block is decided by cross-validation, and later we will also empirically compare different ways of extending sharp attention blocks at different levels of residual blocks.

The optional context-aware unit in the sharp attention block is composed of a downsampled residual unit (whose architecture is reused from the backbone network but with a different group of parameters) and a deconvolutional layer with a kernel filter of size $1 \times 1$ and a fractional $1/2$ stride.

\textbf{Network training.}
The model is implemented on PyTorch and runs on a workstation configured with NVIDIA M40 GPU cards.
For all experiments, within each mini-batch, we randomly sample 64 identities and then randomly sample four images for each person, thus resulting in a batch of $256$ images. We select the hardest positive and the hardest negative samples within a batch to form the triplet units~\cite{hermans2017defense}. We use the pretrained model on ImageNet~\cite{imagenet_cvpr09} to initialize the weights and train the network for up to 150 epochs. Adam optimizer~\cite{kingma2015adam} is adopted to perform weight updates.
We set the initial learning rate to 0.0002 (annealed strategy is according to~\cite{hermans2017defense}), weight decay to 5e-4, triplet margin to 0.5, hyperparameter $\mu$ in loss function Eq.~\ref{eq:8} to 0.1. 
The temperature $\tau$ in Eq.~\ref{eq:5} is initialized to 1.0 and anneals by:

\begin{equation}\label{eq:9}
\tau=
\begin{cases}
1.0 \cdot e^{-\alpha t}, \ \ \ \ \text{if $\tau > \tau_1$ }\\
\tau_1, \ \ \ \ \ \ \ \ \ \ \ \ \ \text{if $\tau \le \tau_1$}
\end{cases},
\end{equation}
\noindent where $\alpha$ is the annealed rate of 0.008, $\tau_1$ denotes the final small temperature of 0.5 and $t$ indexes the epoch.

\subsection{Empirical Analysis}
\label{sec:4_3}

\textbf{The impact of different combinations of sharp attention blocks.} We empirically study different combinations of sharp attention blocks for person re-ID. We conduct an experiment on CUHK03 labeled dataset. The relative performances in terms of mAP (as well as CMC top-1 accuracy) over the ResNet-50 baseline are illustrated in Fig.~\ref{fig:5}(a), (b). Seven different combination strategies, including applying sharp attention to a single block (from block1 to block4), two blocks (block1+block2), three blocks (block1+block2+block3), and even all four blocks, are evaluated in the same experimental setting: the same hyper-parameters and the same sharp attention block design (SAB for short). For the latter, we utilize the proposed sharp attention mask generator (SAMG for short) combined with cross-feature interaction learning (CIL for short), while not involving the optional context-aware unit (CU for short), \ie, \textit{SAMG + CIL, no CU}.

It can be seen that, directly applying Gumbel-Softmax sampling to residual block1 (\ie, low-level feature maps) contributes the major improvement. Involving deep blocks alone, may lead to a slight decline in mAP and CMC top-1 accuracy (\eg, block4). We conjecture the reason is that low-level feature maps (such as block1) contain more subtle visual structures from which the sharp attention selectors can sample discriminative features to uniquely identify different pedestrians. On the contrary, the high-level feature maps (such as block4) are usually sparse and too coarse to distinguish between different people, making selecting discriminative structures from high-level residual block alone less effective than that from low-level one. Further, combining block1 with all other blocks achieves the best performance: an absolute 2.9\% improvement in mAP and 2.7\% improvement in CMC top-1 accuracy over the ResNet-50 baseline. Thus, in all of the next experiments, we adopt such sharp attention block combination strategy. Although applying Gumbel-Softmax sampling to block4 alone leads to a slight decline (just about -0.1\%), the performance of block(1+2+3+4) is a little better than block(1+2+3). We think the main reason is that
the performance gain among different blocks should not be considered as a simple linear additive relationship, because different-level attentions (from different blocks) should have complementary information such that our block-wise (multiple levels of attention learning) design can provide additional top-down attention refinement to boost the re-ID performance collaboratively.

\textbf{Whether or not involving context-aware unit.}
We evaluate how context-aware unit affect the re-ID performance on three datasets: CUHK03 labeled, CUHK03 detected, and Market-1501. Experiments are conducted in the same setting: the same hyper-parameters, and the same sharp attention block design. For the latter, all four residual blocks expand to sharp attention blocks (as clarified above) with SAMG and CIL. The only difference is whether involving the context-aware unit or not. Notice that the context-aware unit should be only added to block1 -- block3, since block4 is the highest-level features already.
	
The experimental results in Fig.~\ref{fig:5}(c) and the comparison between Row-3 and Row-5 of Table~\ref{tab:1} show that using the context-aware unit can consistently improve the performance. 
Specifically, the improvement on CUHK03 labeled is slight， while the effects on CUHK03 detected and Market-1501 are especially significant. These results show that for CUHK03 labeled dataset, the proposed sharp attention mechanism can already obtain pleasing performance without the context-aware unit. However, for CUHK03 detected and Market-1501 datasets, the context-aware unit is essential to play a role of appropriate sampling guiding and cooperate with the sharp attention mechanism for the purpose of acquiring satisfactory results. We think the reason as follows. Compared with CUHK03 labeled dataset where bounding boxes are carefully annotated by human, the other two datasets use DPM detector to produce bounding boxes of people. 
Thus, the latter two datasets are prone to misalignment 
due to the poor detected bounding boxes.
They represent more challenging scenarios than CUHK03 labeled dataset. Therefore, involving a larger receptive field and high-level information in context-aware unit could be helpful to alleviate the negative impact from background clutter, occlusion and missing body parts, and supply adequate guidance to more accurately generate attentions on identifying different pedestrians. Notice that, we do not imply that the context-aware unit is the largest contributor in most cases, since context-aware unit is just a front-end tool for sampling guiding and cannot be utilized solely. Instead, the proper statement is that the context-aware unit is essential and fundamental to assist the proposed sharp attention achieving significant effects for those more challenging re-ID scenarios.


\begin{table*}[t]
	\centering
	\caption{Performance comparison of different sharp attention block designs (SAB, \ie, different components and their combinations) on several datasets. The CMC rank-1 accuracy (\%) and mAP (\%) are presented. ``SAMG" means sharp attention mask generator, ``CIL" means cross-feature interaction learning, 
		``CU" means context-aware unit, ``TV" means TV regularization penalty.}
	\label{tab:1}
	\scalebox{1.1}{
		\begin{tabular}{l|cccccccccccc}
			\hline
			\multicolumn{1}{c|}{\multirow{2}{*}{Method}} &           & \multicolumn{2}{l}{CUHK03  labeled} &           & \multicolumn{2}{l}{CUHK03 detected} &           & \multicolumn{2}{l}{Market-1501}  &       & 
			\multicolumn{2}{l}{DukeMTMC-reID}    \\ \cline{3-4} \cline{6-7} \cline{9-10} \cline{12-13}
			\multicolumn{1}{c|}{}                        &           & rank-1            & mAP              &           & rank-1            & mAP              &           & rank-1          & mAP  &           & rank-1          & mAP          \\ \hline
			ResNet-50 baseline                            &           & 85.1              & 82.1             &           & 82.1              & 79.7             &           & 84.0            & 67.9   &           & 75.3            & 56.4         \\ 
			Baseline + SAMG                        &           & 87.2              & 84.4             &           & -              & -             &           & -            & -    &           & -            & -       \\ 
			Baseline + SAMG + CIL                           &           & 87.8              & 85.0             &           & 82.5              & 80.3             &           & 84.3            & 68.3     &           & 75.9            & 57.1       \\
			Baseline + SAMG + CU                    &           & -              & -             &           & 83.4              & 81.2             &           & 85.2            & 69.0    &           & 76.9            & 57.9        \\  
			Baseline + SAMG + CIL + CU                   &           & 88.0              & 85.4             &           & 83.9              & 81.6             &           & 85.6            & 69.6    &           & 77.5            & 58.4        \\ 
			\textbf{Baseline + SAMG + CIL + CU + TV}             & \textbf{} & \textbf{88.3}     & \textbf{85.9}    & \textbf{} & \textbf{84.3}     & \textbf{82.2}    & \textbf{} & \textbf{85.9}   & \textbf{70.1}  & \textbf{} & \textbf{77.9}   & \textbf{58.8}  \\ \hline
	\end{tabular}}
\end{table*}


\textbf{The separate effectiveness of sharp attention mask generator (SAMG) and cross-feature interaction learning (CIL).} 
Moreover, SAMG is evaluated with/without CIL to demonstrate their individual effectiveness. As stated in the previous paragraph, for CUHK03 detected, Market-1501 and DukeMTMC-reID datasets, we conduct experiments under the circumstance of involving context-aware unit, since context-aware unit is essential to guarantee the sharp attention mechanism making effects for those more challenging re-ID scenarios. While for CUHK03 labeled dataset, context-aware unit is not required. As can be seen from the comparisons of Row-2, Row-3 of Table~\ref{tab:1} (for CUHK03 labeled) and  Row-4, Row-5 of Table~\ref{tab:1} (for the other three datasets), SAMG plays a major role and 
containing CIL further improves the re-ID performance consistently. Therefore, these two  are both effective components among our sharp attention block design.

\textbf{The effectiveness of optimizing attention-aware feature continuity.} 
On the basis of the above optimal experimental practice (all four blocks combination: SAMG + CIL and involving CU), we further add the TV regularization penalty, aiming to eliminate noisy features and optimize the continuity of attention-aware features. The results between Row-5 and Row-6 of Table~\ref{tab:1} clearly present a slight performance gain, about +0.3\% -- +0.6\% for all datasets, demonstrating the effectiveness of our introduced regularization technique.

\subsection{Comparison with the ResNet-50 Baseline}
\label{sec:4_4}

In this subsection, the overall performances of the proposed method are summarized (ablation study has been demonstrated in the previous subsection). We adopt ResNet-50~\cite{He_2016_CVPR} network trained with triplet loss (using the triplet mining strategy introduced by Hermans \etal~\cite{hermans2017defense}) to obtain a strong CNN baseline for all datasets. As shown in Table~\ref{tab:1}, we have the baseline results as follows. The mAP is 82.1\%, 79.7\%, 67.9\%, and 56.4\% on CUHK03 labeled, CUHK03 detected, Market-1501 and DukeMTMC-reID datasets, respectively. The corresponding CMC rank-1 accuracy is 85.1\%, 82.1\%, 84.0\% and 75.3\%. Note that, the baseline alone exceeds most of the existing deep learning frameworks (see Table~\ref{tab:3} -- Table~\ref{tab:6} in detail).
We adopt the practice discussed in Sec.~\ref{sec:4_3} to validate the effectiveness of the proposed sampling-based sharp attention mechanism. Table~\ref{tab:1} detailed summarizes the positive effects of the proposed algorithm modules over the baseline, including the basic sharp attention block design (\ie, all four residual blocks expanding to sharp attention blocks with sharp attention mask generator and cross-feature interaction learning, refer to Sec.~\ref{sec:3_1} and Sec.~\ref{sec:3_3}), the optional context-aware unit (refer to Sec.~\ref{sec:3_4}) and the TV regularization optimization (refer to Sec.~\ref{sec:3_2}).
The results show that we gain significant improvements consistently in both mAP and CMC rank-1 accuracy over the strong baseline on all datasets. Specifically, we observe final improvements of +3.8\%, +2.5\%, +2.2\%, +2.4\% in mAP and +3.2\%, +2.2\%, +1.9\%, +2.6\% in CMC rank-1 accuracy. These results also 
demonstrate that all four core components in our approach are pretty important designs, improving the baseline and achieving splendid performances in various scenarios steadily and consistently.

\begin{table*}[t]
	\centering
	\caption{Comparisons of additional parameter number and performance on CUHK03 labeled dataset. The additional parameter number (million) indicates the extra parameters needed in the context-aware unit. For performance comparison, the CMC rank-1 accuracy (\%) and mAP (\%) are listed.}
	\label{tab:2}
	\scalebox{1.2}{
		\begin{tabular}{l|cccccccc}
			\hline
			\multicolumn{1}{c|}{\multirow{2}{*}{Method}} &  & \multicolumn{4}{l}{Additional parameter number(M)}                                                                            &  & \multicolumn{2}{l}{CUHK03 labeled} \\ \cline{3-6} \cline{8-9}
			\multicolumn{1}{c|}{}                        &  & \multicolumn{1}{c}{block1}        & \multicolumn{1}{c}{block2}        & \multicolumn{1}{c}{block3}        & total          &  & rank-1           & mAP              \\ \hline
			Soft attention network~\cite{Wang_2017_CVPR}                    &  & \multicolumn{1}{c}{15.84}         & \multicolumn{1}{c}{15.08}         & \multicolumn{1}{c}{12.06}         & 42.98          &  & 87.0             & 85.1             \\ \hline
			\textbf{Our sharp attention network}          &  & \multicolumn{1}{c}{\textbf{0.51}} & \multicolumn{1}{c}{\textbf{2.03}} & \multicolumn{1}{c}{\textbf{8.13}} & \textbf{10.67} &  & \textbf{88.3}    & \textbf{85.9}    \\ \hline
	\end{tabular}}
\end{table*}

\begin{figure}[]
	\begin{center}
		\includegraphics[width=0.8\linewidth,height=7cm]{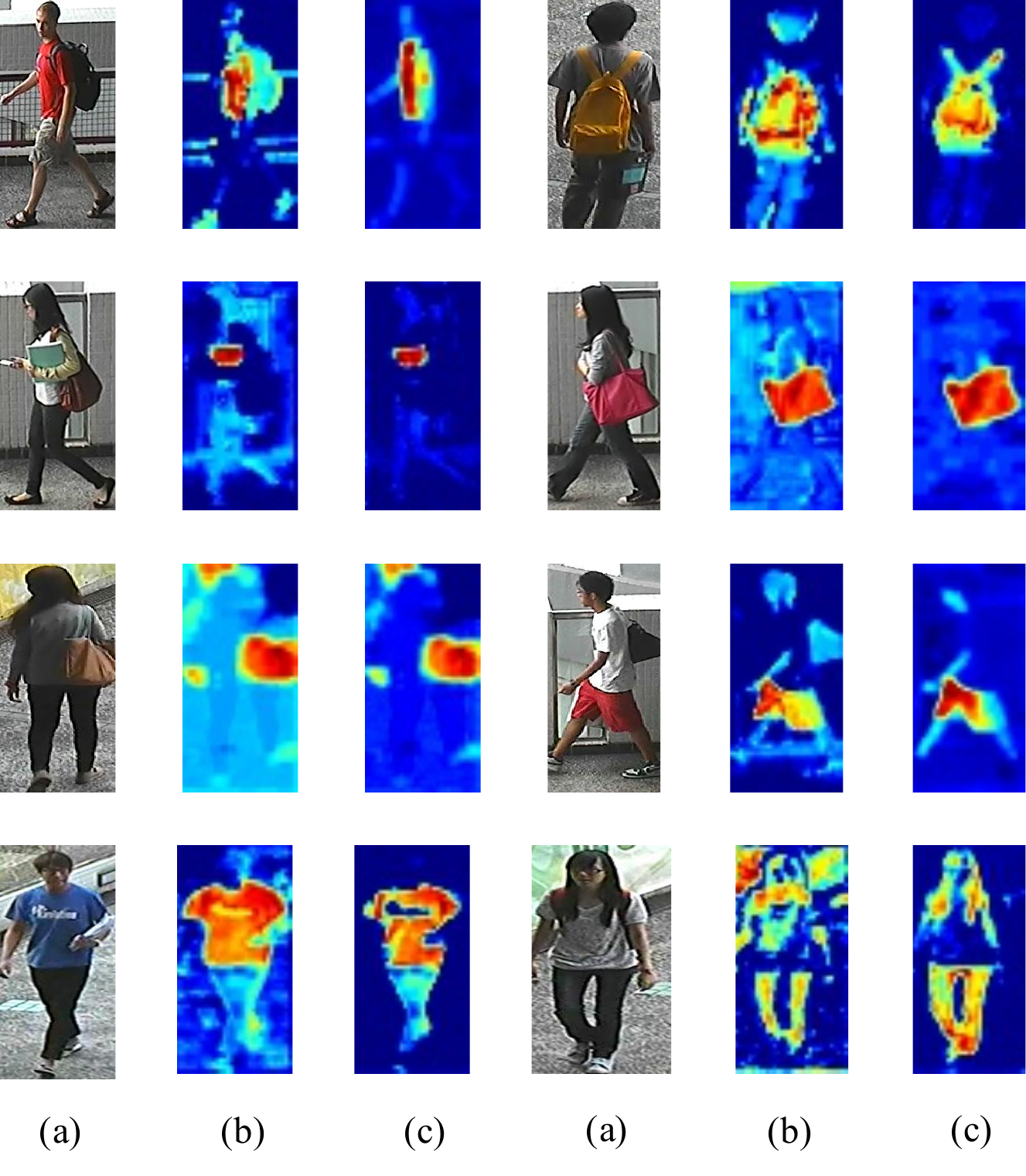}
	\end{center}
	\caption{Visualization of different types of Conv1 attention features. (a) Raw images. (b) Gating-based soft attention feature maps. (c) Sampling-based sharp attention feature maps. These qualitative results indicate the proposed  sharp attention mechanism is more assertive in localizing subtle visual details.}
	\label{fig:6}
\end{figure}

\subsection{Comparison with Soft Attention Models}
\label{sec:4_5}

\cite{rahimpour2017person, liu2017end, Zhao_2017_ICCV, Wang_2017_CVPR} represent four state-of-the-art soft attention models, from which we firstly choose the method proposed by~\cite{Wang_2017_CVPR} to compare its performance with ours. We highlight the comparison with~\cite{Wang_2017_CVPR} because it performs best among the above-mentioned four methods (see Table~\ref{tab:3} -- Table~\ref{tab:5}). Another reason is that it is easily compared with us, on the contrary, other methods~\cite{rahimpour2017person, liu2017end, Zhao_2017_ICCV} contain some additional techniques (such as RNN, STN~\cite{jaderberg2015spatial}), which are not directly comparable to us. For direct and fair comparison between sharp attention and soft attention, we conduct experiments within the same setting: the same backbone network (\ie, ResNet-50), the same loss function (\ie, triplet loss), the same hyper-parameters and the same add-ons (\ie, CIL and CU). The major difference is that our method employ a distinguished sampling-based attention generation mechanism rather than soft attention ones. Besides, the context-aware unit structure utilized by us is much more concise.

According to~\cite{Wang_2017_CVPR}, we reproduce the proposed residual attention network based on ResNet-50 backbone and conduct experiments on CUHK03 labeled dataset. Results from Table~\ref{tab:2} show that we only use 25\% parameter number (10.67M with 42.98M) within context-aware unit while outperforms by +1.3\% (88.3\% with 87.0\%) in CMC rank-1 accuracy and +0.8\% (85.9\% with 85.1\%) in mAP, comparing to~\cite{Wang_2017_CVPR}. That is attribute to the superiority of sharper attention selectors over soft attention ones. 
For some typical difficult cases intuitively reflected in Fig.~\ref{fig:6} and the former Fig.~\ref{fig:2}, 
the soft attentions look ambiguous in selecting subtle discriminative parts, because their mask values are far from two assertive statuses of being attended ($1$) or unattended ($0$) (shown in Fig.~\ref{fig:2}(b)). Compared with the soft attentions, the sharp attentions are more certain and assertive in selecting discriminative visual structures (\eg, the regions of backpack, T-shirt, book or pants, shown in Fig.~\ref{fig:6}) to identify people, which makes them particularly suitable for solving person re-ID in more challenging scenarios. Also thanks to the discrete nature of sharp attention samplers which are more aggressive and assertive on selecting attended/unattended visual structures, we just need a relatively compact front-end unit (\ie, less parameters) for visual contexts modeling and attention generation guiding. This makes our network more effective in model deployment and reduces the risk of overfitting.

More experimental comparison results with other soft attention methods on other datasets are detailed  summarized in Table~\ref{tab:3} -- Table~\ref{tab:5}. For~\cite{liu2017end, rahimpour2017person, Zhao_2017_ICCV}, we just list the results reported in their papers. We find that the proposed method outperforms the soft attention models 
on all datasets, including CUHK03 labeled, CUHK03 detected and Market-1501. This demonstrates that, our approach not only provides an alternative way to generate attentions, but also achieves the state-of-the-art performance for person re-ID within different attention models.

\begin{table}[t]
	\centering
	\caption{Performance comparison with other simple alternatives on CUHK03 labeled and Market-1501 datasets. The CMC rank-1 accuracy (\%) and mAP (\%) are presented.}
	\label{tab:7}
	\begin{tabular}{l|cccccc}
			\hline
			\multicolumn{1}{c|}{\multirow{2}{*}{Method}} &           & \multicolumn{2}{l}{CUHK03  labeled} &           &  \multicolumn{2}{l}{Market-1501}    \\ \cline{3-4} \cline{6-7} 
			\multicolumn{1}{c|}{}                        &           & rank-1            & mAP              &           & rank-1            & mAP            \\ \hline
			ResNet-50 baseline & & 85.1 & 82.1 & & 84.0  & 67.9  \\ 
			Thresholding  &  & 85.3  & 82.4  &  & 84.1   & 68.0   \\ 
			Power ($x^2$)  &   & 86.1   & 83.1 &    & 84.4  & 68.2  \\
			Power ($x^3$)  &  & 85.8  & 82.7  &  & 84.3  & 68.1       \\  
			\textbf{Our SAN}    &   & \textbf{87.2}    & \textbf{84.4}             &        & \textbf{85.2}   & \textbf{69.0}     \\  \hline
	\end{tabular}
\end{table}

\subsection{Comparison with Other Simple Alternatives}
\label{sec:4_6}

As mentioned in Sec.~\ref{sec:3_1}, the normalized feature Eq.~\ref{eq:3}
is viewed as the probability of sampling this feature, which aims to generate attention-aware features in which strong gets stronger, and weak becomes weaker or even vanishes. We evaluate the proposed sampling-based mechanism with other two simple alternatives which share the similar intuition, on CUHK03 labeled and Market-1501. The first is thresholding, where features that are below a certain threshold $\lambda$ are set to 0, and the second is power (\eg, square or cube). We tune the hyperparameter $\lambda$ to 0.3 on the validation dataset. Experiments are conducted in the same setting (without CIL and TV, CU are only involved for Market-1501). The only difference lies in the attention mask generation techniques.
The results of Table~\ref{tab:7} show that our SAMG outperforms other simple  alternatives, which demonstrates the effectiveness of our delicately designed sharp attention mechanism.

\begin{table}
	\centering
	\caption{Performance comparison on CUHK03 labeled dataset. The compared methods are separated into two categories: gating-based soft attention models (GSA) and other state-of-arts (SOA). The CMC rank-1/5/10 accuracy (\%) and mAP (\%) are presented.}
	\label{tab:3} 
	\small
	\setlength\tabcolsep{5pt}
	\begin{tabular}{l|l|cccc}
		\hline
		\multicolumn{2}{c|}{Method}                                                                            & rank-1        & rank-5        & rank-10       & mAP  \\ \hline
		\multicolumn{1}{c|}{\multirow{4}{*}{\begin{tabular}[c]{@{}c@{}}GSA\end{tabular}}} & GAN~\cite{rahimpour2017person}          & 61.2          & 89.1          & 91.3          & -    \\ 
		\multicolumn{1}{c|}{}                                                                   & CAN~\cite{liu2017end}          & 77.6          & 95.2          & 99.3          & -    \\ 
		\multicolumn{1}{c|}{}                                                                   & Part Aligned~\cite{Zhao_2017_ICCV} & 85.4          & 97.6          & 99.4          & -    \\ 
		\multicolumn{1}{c|}{}                                                                   & SAN~\cite{Wang_2017_CVPR} & 87.0         & 98.3          & 99.4          & 85.1    \\ \hline
		\multirow{8}{*}{\begin{tabular}[c]{@{}l@{}}SOA\end{tabular}}                   & LOMO~\cite{Liao_2015_CVPR}    & 52.2          & 82.2          & 92.1          & -    \\ 
		& DNS~\cite{Zhang_2016_CVPR}          & 58.9          & 85.6          & 92.5          & -    \\ 
		& Latent Part~\cite{Li_2017_CVPR}  & 74.2          & 94.3          & 97.5          & -    \\  
		& SSM~\cite{Bai_2017_CVPR}          & 76.6          & 94.6          & 98.0          & -    \\  
		& MuDeep~\cite{Qian_2017_ICCV}       & 76.9          & 96.1          & 98.4          & -    \\  
		& MSP-CNN~\cite{Shen2017Deep}      & 85.7          & 97.6          & 99.2          & -    \\ 
		& Spindle~\cite{Zhao_2017_CVPR}      & 88.5          & 97.8          & 98.6          & -    \\  
		& PDC~\cite{Su_2017_ICCV}          & \textbf{88.7} & 98.6          & 99.2          & -    \\ \hline
		& \textbf{Our SAN}      & 88.3          & \textbf{98.8} & \textbf{99.4} & \textbf{85.9} \\
		\hline
	\end{tabular}
\end{table}



\begin{table}
	\centering
	\caption{Performance comparison on CUHK03 detected dataset. The CMC rank-1/5/10 accuracy (\%) and mAP (\%) are shown.}
	\label{tab:4} 
	\small
	\setlength\tabcolsep{5pt}
	\begin{tabular}{l|l|cccc}
		\hline
		\multicolumn{2}{c|}{Method}                                  & rank-1        & rank-5        & rank-10       & mAP           \\ \hline
		\multicolumn{1}{c|}{\multirow{3}{*}{GSA}} & CAN~\cite{liu2017end}              & 69.2          & 88.5          & 94.1          & -             \\  
		\multicolumn{1}{c|}{}                     & Part-Aligned~\cite{Zhao_2017_ICCV}     & 81.6          & 97.3          & 98.4          & -             \\ 
		\multicolumn{1}{c|}{}                     & 
		SAN~\cite{Wang_2017_CVPR}     & 83.1          & 97.3          & 98.3          & 81.0             \\ \hline
		\multirow{9}{*}{SOA}                       & LOMO~\cite{Liao_2015_CVPR}             & 46.3          & 78.9          & 88.6          & -             \\  
		& SI-CI~\cite{Wang_2016_CVPR}            & 52.2          & 84.3          & 94.8          &               \\ 
		& DNS~\cite{Zhang_2016_CVPR}              & 53.7          & 83.1          & 93.0          & -             \\  
		& Latent Part~\cite{Li_2017_CVPR}      & 68.0          & 91.0          & 95.4          & -             \\  
		& SSM~\cite{Bai_2017_CVPR}              & 72.7          & 92.4          & 96.1          & -             \\  
		& LSRO~\cite{Zheng_2017_ICCV}             & 73.1          & 92.7          & 96.7          & 77.4          \\  
		& MuDeep~\cite{Qian_2017_ICCV}           & 75.6          & 94.4          & 97.5          & -             \\  
		& PDC~\cite{Su_2017_ICCV}              & 78.3          & 94.9          & 97.2          & -             \\  
		& SVDNet~\cite{Sun_2017_ICCV}           & 81.8          & -             & -             & -             \\ \hline
		& \textbf{Our SAN} & \textbf{84.3} & \textbf{97.4} & \textbf{98.4} & \textbf{82.2} \\
		\hline
	\end{tabular}
\end{table}




\subsection{Comparison with the State-of-the-Arts}
\label{sec:4_7}

Finally, we compare our method with the existing published state-of-the-art methods on CUHK03 (including both labeled and detected settings), Market-1501 and  DukeMTMC-reID in Table~\ref{tab:3} -- Table~\ref{tab:6}, respectively. On CUHK03 labeled dataset, we have 88.3\% in rank-1 accuracy, 98.8\% in rank-5 accuracy, and 99.4\% in rank-10 accuracy. The rank-5 and rank-10 accuracy reach the best results in literature. We also achieve the mAP of 85.9\%, which is also very competitive. On CUHK03 detected dataset, we achieve the best results: 84.3\% in CMC rank-1 accuracy and 82.2\% in mAP. On Market-1501 dataset, we also beat all other methods with 85.9\% in rank-1 accuracy and 70.1\% in mAP, even some models (\eg, the second best one~\cite{Bai_2017_CVPR}) utilize an additional re-ranking technique. For the newest DukeMTMC-reID dataset, we reach a new state-of-the-art performance as well: 77.9\% in rank-1 accuracy, 58.8\% in mAP.
All these results demonstrate the superiority of our novel sharp attention model for person re-ID over the other existing published state-of-the-art approaches.

\begin{table}[t]
	\centering
	\caption{Performance comparison on Market-1501 dataset. The CMC rank-1/5/10 accuracy (\%) and mAP (\%) are shown.}
	\label{tab:5}
	\small
	\setlength\tabcolsep{5pt}
	\begin{tabular}{l|l|cccc}
		\hline
		\multicolumn{2}{c|}{Method}                                  & rank-1                    & rank-5                    & rank-10                   & mAP                       \\ \hline
		\multicolumn{1}{c|}{\multirow{3}{*}{GSA}} & CAN~\cite{liu2017end}              & 60.3                      & -                         & -                         & 35.9                      \\  
		\multicolumn{1}{c|}{}                     & Part-Aligned~\cite{Zhao_2017_ICCV}     & 81.0                      & 92.0                      & 94.7                      & 63.4                      \\ 
		\multicolumn{1}{c|}{}                     & 
		SAN~\cite{Wang_2017_CVPR}     & 84.8                      & 94.2                      & 96.7                      & 69.2                     \\ \hline
		\multirow{8}{*}{SOA}                       & DNS~\cite{Zhang_2016_CVPR}              & 55.4                      & -                         & -                         & 29.9                      \\  
		& Spindle~\cite{Zhao_2017_CVPR}          & 76.9                      & 91.5                      & 94.6                      & -                         \\  
		& LSRO~\cite{Zheng_2017_ICCV}             & 78.1                      & -                         & -                         & 56.2                      \\  
		& Latent Part~\cite{Li_2017_CVPR}      & 80.3                      & -                         & -                         & 57.5                      \\  
		& MSP-CNN~\cite{Shen2017Deep}          & 81.9                      & 92.8                      & 95.2                      & 63.6                      \\  
		& SVDNet~\cite{Sun_2017_ICCV}           & 82.3                      & -                         & -                         & 62.1                      \\  
		& SSM~\cite{Bai_2017_CVPR}              & 82.2                      & -                         & -                         & 68.8             \\  
		& PDC~\cite{Su_2017_ICCV}              & \multicolumn{1}{c}{84.4} & \multicolumn{1}{c}{92.7} & \multicolumn{1}{c}{94.9} & \multicolumn{1}{c}{63.4} \\ \hline
		& \textbf{Our SAN} & \textbf{85.9}             & \textbf{94.9}             & \textbf{97.0}             & \textbf{70.1}  \\
		\hline                 
	\end{tabular}
\end{table}




\section{Conclusion}
\label{sec:5}

In this paper, we propose a sampling-based sharp attention mechanism, which can generate sharper attention masks that are more assertive on selecting discriminative visual structures than gating-based soft attention models by directly sampling from the convolutional features. The sharper attention mask is adequate to distinguish subtle visual details from irrelevant parts, and is particularly suitable for solving challenging recognition problems like person re-ID.
A differentiable Gumbel-Softmax sampler is employed to approximate the Bernoulli sampling to train the sharp attention networks in a end-to-end fashion. We further introduce a compact context-aware unit to capture high-level context-aware features to better guide sampling of attention masks in complex contexts.
Experiments on three large-scale datasets demonstrates the superiority of the proposed approach over gating-based soft attention models as well as  other existing published state-of-the-art methods.






\begin{table}
	\centering
	\caption{Performance comparison on DukeMTMC-reID dataset. The CMC rank-1 accuracy (\%) and mAP (\%) are listed.}
	\label{tab:6}
	\small
	\setlength\tabcolsep{12pt}
	\begin{tabular}{l|cc}
		\hline
		\multicolumn{1}{c|}{Method} & rank-1        & mAP           \\ \hline
		LOMO+XQDA~\cite{Liao_2015_CVPR}                   & 30.8          & 17.0          \\ 
		LSRO~\cite{Zheng_2017_ICCV}                         & 67.7          & 47.1          \\ 
		OIM ~\cite{Xiao_2017_CVPR}                         & 68.1          & -             \\ 
		ACRN~\cite{Arne_2017_CVPR_Workshops}                         & 72.6          & 52.0          \\ 
		SVDNet~\cite{Sun_2017_ICCV}                       & 76.7 & 56.8 \\ \hline
		\textbf{Our SAN}             & \textbf{77.9}          & \textbf{58.8}          \\ \hline
	\end{tabular}
\end{table}

\appendices


\section*{Acknowledgment}
This work was supported in part by the Fundamental Research Funds for the Central Universities.


\ifCLASSOPTIONcaptionsoff
  \newpage
\fi




\bibliographystyle{IEEEtran}
\bibliography{T-CSVT_sharp_attention_network}
\end{document}